\pdfoutput=1
\documentclass[11pt]{article}

\usepackage{acl}

\usepackage{times}
\usepackage{latexsym}

\usepackage[T1]{fontenc}
\usepackage[utf8]{inputenc}

\usepackage{microtype}

\usepackage{array}
\usepackage{float}
\usepackage{arydshln}
\usepackage{booktabs}
\usepackage{pdfpages}
\usepackage{hyperref}
\usepackage{multirow}
\usepackage{tikz}
\usepackage{xcolor}
\usepackage{dsfont}
\usepackage[ruled,linesnumbered]{algorithm2e}
\usepackage{placeins}
\usepackage{amsmath}

\let\oldnl\nl% Store \nl in \oldnl
\newcommand{\nonl}{\renewcommand{\nl}{\let\nl\oldnl}}% Remove line number for one line

\newcommand{\rectangle}[1]{
\begin{tikz}
\draw [fill=#1,#1] rectangle (0.05,0.24);
\end{tikz}
}

\definecolor{gold}{HTML}{FFD700}
\definecolor{lime}{HTML}{00FF00}
\definecolor{cyan}{HTML}{00FFFF}
\definecolor{fuchsia}{HTML}{FF00FF}
\definecolor{darkgreen}{HTML}{006400}
\definecolor{limegreen}{HTML}{32CD32}
\definecolor{greenyellow}{HTML}{ADFF2F}
\definecolor{darkviolet}{HTML}{9400D3}
\definecolor{lightpink}{HTML}{FFB6C1}

\newcolumntype{A}{p{0.04\textwidth}}
\newcolumntype{B}{p{0.05\textwidth}}
\newcolumntype{C}{p{0.06\textwidth}}

\title{Can the Variation of Model Weights be used as a\\Criterion for Self-Paced Multilingual NMT?}

\author{\`{A}lex R. Atrio$^{1,2}$, Alexis Allemann$^{1}$, Ljiljana Dolamic$^{3}$ \and Andrei Popescu-Belis$^{1,2}$ \\[8pt] 
\begin{tabular}{c}
$^1$HEIG-VD / HES-SO, Yverdon-les-Bains, Switzerland \\
$^2$EPFL, Lausanne, Switzerland \\
$^3$Armasuisse W+T, Thun, Switzerland \\
\texttt{\footnotesize name.surname@heig-vd.ch, ljiljana.dolamic@armasuisse.ch}\\
\end{tabular}
}

\begin{document}
\maketitle
\begin{abstract}
Many-to-one neural machine translation systems improve over one-to-one systems when training data is scarce.  In this paper, we design and test a novel algorithm for selecting the language of minibatches when training such systems.  The algorithm changes the language of the minibatch when the weights of the model do not evolve significantly, as measured by the smoothed KL divergence between all layers of the Transformer network.  This algorithm outperforms the use of alternating monolingual batches, but not the use of shuffled batches, in terms of translation quality (measured with BLEU and COMET) and convergence speed.
\end{abstract}

% = = = = = = = = = = = = = = = = = = = = = = = = = = =
\section{Introduction}
\label{sec:introduction}

Multilingual neural machine translation (MNMT) systems can be trained with several languages on the source side, or on the target side, or on both sides \citep{firat-etal-2016-multi,johnson-etal-2017-googles}.  Many-to-one MNMT systems are particularly effective for low-resource languages (LRLs) on the source side, when they are accompanied by high-resource languages (HRLs) related to them \citep{gu-etal-2018-universal}.  For instance, \citet{neubig-hu-2018-rapid} trained a many-to-one recurrent model on a multilingual dataset of almost 60 languages and showed that including HRLs in the training data reduces the chance of overfitting to the LRLs and improves translation quality. \citet{aharoni-etal-2019-massively} used Transformer models \citep{NIPS2017_3f5ee243} to further improve over these results.

Many-to-one MNMT systems are usually trained with multilingual batches sampled from all source languages to avoid catastrophic forgetting \citep{jean2019adaptive}, but the presence of several languages in a minibatch may ineffectively constrain the model and prevent it from training on the languages where training is most needed.  An open question in many-to-one MNMT, therefore, is how the data from different source languages should be sampled during training, particularly when massive imbalances in sizes or difficulties occur across languages.

In this paper, we propose a dynamic scheduling approach which samples minibatches from the source languages based on the variation of weights in the layers of a Transformer.  The main idea is the following one: when a model becomes competent for translating a certain source language, as indicated by a decreasing variation of a model's weights across training steps, then the language of the minibatches should be switched to a new one, in order to allocate more time to more challenging, hence useful tasks.  

% We propose to train on monolingual batches (i.e., all the samples on the source side of a batch are of the same language), and use the model's self-assessed competence on each language to optimize on which language it should train at each step.
The main contributions of the paper are the precise formulation, implementation and testing of the idea.  Specifically, we propose to:
\begin{itemize} \setlength{\itemsep}{0pt}
    \item measure variation of weights by comparing the weights of all layers of a Transformer across two consecutive training steps with the same source language;
    \item compare weights by using symmetric KL divergence between softmaxes of layers, with exponential smoothing across time;
    \item trigger a change of task, i.e. source language, when weight variation decreases;
    \item compare translation quality and convergence speed for 8-to-1 MNMT on a dataset with four language families on the source side, and one HRL and one LRL for each of them \citep{neubig-hu-2018-rapid}.
\end{itemize}

Although experimental results suggest that a simple random task mixing approach is sufficient to optimize model performance, the paper presents various techniques to better understand a system's behavior during learning.

% = = = = = = = = = = = = = = = = = = = = = = = = = = =
\section{Related Work}
\label{sec:sota}

% multilingual NMT
\citet{neubig-hu-2018-rapid} study the upsampling of the HRL data when building minibatches, and observe that keeping the original proportions of HRL and LRL performs marginally better.
\citet{aharoni-etal-2019-massively} also sample each batch uniformly from a concatenation of all language pairs.
\citet{arivazhagan2019massively} compare a simple concatenation with uniform balancing \citep{johnson-etal-2017-googles}, but observe better results for LRLs when translating into a HRL by using a temperature-based upsampling, which has been favored afterwards \citep{conneau-etal-2020-unsupervised,tang-etal-2021-multilingual}.

% self-pacing / competence
As a method for dynamic scheduling of multitask training \citep{caruana1997multitask}, self-pacing consists in using the target model to quantify the difficulty of each sample or dataset -- that is,  measure the model's \emph{competence} -- and inform the scheduling module dynamically \citep{kumar2010self}. Self-pacing has been used in NMT at the sample-level, for instance by measuring variance across dropout runs \citep{wan-etal-2020-self}. Similarly, \citet{liu-etal-2020-norm} set a self-paced curriculum based on the norm of a token's embedding, for a single task.

% adaptive MNMT
For MNMT, \citet{jean2019adaptive} compare adaptively upsampling a language depending on various factors, observing best results on the LRLs when dynamically changing the gradient norm \citep{chen2018gradnorm}.
\citet{wang-etal-2020-balancing} adaptively balance the languages by learning language weights on the model's competence on a development set.
\citet{zhang-etal-2021-competence-based} adaptively learn a sampling strategy by measuring per-language competence and LRL competence evaluated with a HRL's competence.
\citet{wu-etal-2021-uncertainty} also balance the data dynamically by measuring the model's uncertainty on a development set, estimated by the variance over several runs of Monte Carlo dropout \citep{pmlr-v48-gal16}.  For one-to-many MNMT, \citet{wang-etal-2018-three} proposed three solutions to modify the decoder in order to improve performance.

% = = = = = = = = = = = = = = = = = = = = = = = = = = =
\section{Method for Self-Paced MNMT}
\label{sec:self-paced}

\subsection{Formulation and Implementation}

To train a many-to-one MNMT model, we consider $M$ parallel datasets which correspond to as many tasks $\mathcal{T} = \{T_1, \ldots, T_M\}$ with different source languages and their respective English translations. Our algorithm chooses on which task $T_c$ to train the MNMT model with parameters $\theta{_t}$ at each time step $t$, based on an estimation of the model's competence for each task (i.e., source language).  The overall goal is to \textit{increase time spent on tasks where the model is less competent, and to avoid over-training on tasks where the model is already competent}. 

We estimate the per-task competence of the model as the average variation of its weights in all layers (due to the back-propagation of gradients) at a given training step.  We thus measure competence as the Kullback-Leibler divergence ($D_\text{KL}$) between the updated weights and the weights at the previous step at which the model was trained on the same task. 

Originally used to quantify the dissimilarity between two probability distributions $P$ and $Q$, $D_\text{KL}$ is defined as: $D_\text{KL}(P||Q)=\sum_{x}P(x)\log(P(x) / Q(x))$ where $x$ are the possible values of the $P$ and $Q$ random variables.  To use $D_\text{KL}$ as a distance measure between two sets of weights in a neural network, we apply softmax $\sigma$ to convert the weights to probability distributions.  Moreover, we take the logarithm of the first term in KL to handle the potential issue of capacity overflow and maintain the stability of divergence calculations \citep{liang-2021-rdrop}.  Finally, we symmetrize the distance by summing KL divergence in both directions.
Therefore, we compute the average variation between two sets of values $\theta_{t-1}$ and $\theta_{t}$ of all the trainable weights of a Transformer network (layers 1 through $L$) as follows:
\begin{equation*}
\begin{aligned}
D(\theta_{t-1},\theta_{t}) = & \frac{1}{2L}\sum_{i=1}^{L}
D_{\text{KL}}(\log(\sigma(\theta_{t-1}^{i})) || \sigma(\theta_{t}^{i}))\ \\
& + D_{\text{KL}}(\log(\sigma(\theta_{t}^{i})) || \sigma(\theta_{t-1}^{i})).
\label{eq:weight-variation}
\end{aligned}
\end{equation*}

Furthermore, we ensure that when the training switches to another task, the model trains on it for at least two updates, so that both $\theta_{t-1}$ and $\theta_t$ are the result of training on minibatches of the same source language: with this, we avoid measuring a large variation between weights simply as the result of switching between tasks. 
In order to obtain a task-switching schedule that is robust to local variations, we apply exponential smoothing and compute per-task competence, transforming $D$ into $D'_c$ as follows: 
\begin{equation*}
D'_c(\theta_{t-1},\theta_{t}) = (1-w) D(\theta_{t-1},\theta_{t}) + w D'_c(\theta_{t-k},\theta_{t-1}),
\end{equation*}
% $D'_c(\theta_{t-1},\theta_{t}) = (1-w) D(\theta_{t-1},\theta_{t}) + w D'_c(\theta_{t-k},\theta_{t-1})$,
where $k \geq 2$ is the smallest value such that $B_{t-k} \in T_c$ (in other words, $t-k$ is the latest step before $t-1$ for which $B_{t-k} \in T_c$). The smoothing weight was set at $w=0.995$ after empirical analyses (see Appendix~\ref{subsec:smoothing}).

The proposed algorithm for dynamic scheduling (Algorithm~\ref{alg:dynamic_sampling} below) has the following rationale.  If the network is trained on a task $T_c$ and the weight variation across consecutive steps increases, we consider that the network lacks competence on $T_c$ and should keep training on it. Conversely, the less the weights change, the more competent the model is.  So, if weight variation slows down, then training on the same task produces diminishing returns, and the network should switch to a task on which it is less competent. This condition appears in line~8 of Algorithm~\ref{alg:dynamic_sampling}.

We define the model's per-task competences at step $t$ as $\mathcal{C}=\{C_1, \ldots, C_M\}$, such that $C_c = D'_c(\theta_{j-1}, \theta_{j})$, and $j \leq t$ is the last step such that minibatch $B_j \in T_c$. That is, for each $T_i \in \mathcal{T}$, $C_i$ is the result of exponential smoothing over the weight variations of all the updates in which $\theta$ has trained on a minibatch from $T_i$.
We define a sampling function -- noted `sample$^*$' in line~10 of the algorithm -- with the following role:
\begin{itemize} \setlength{\itemsep}{0pt}
    \item in the initial phase, it randomly samples any of the $T_i \in \mathcal{T}$ on which the system has never been trained on; 
    \item then, when all tasks have been seen at least once, it samples a new $T_c \in \mathcal{T}$ based on the softmaxed per-task competence distribution $\sigma(\mathcal{C})$.
\end{itemize}
% in the initial phase, it randomly samples any of the $T_i \in \mathcal{T}$ on which the system has never been trained on; then, when all tasks have been seen at least once, it samples a new $T_c \in \mathcal{T}$ based on the softmaxed per-task competence distribution $\sigma(\mathcal{C})$.
% 
Additionally, we introduce hyper-parameter $\alpha$ in order to compare the importance of previous weight variation versus the current one (line 8).  However, after empirical analyses, we found that the best results were obtained with $\alpha=1$ (see Appendix~\ref{subsec:alpha}).

\begin{algorithm}[ht!]
    \SetKwInput{Input}{Input}
    \SetKwInput{Require}{Require}
    \SetKwInOut{Output}{Output}
    \SetAlgoLined
    \Require{tasks $\mathcal{T} = \{T_1, .., T_M\}$, steps $s$}
    $T_c \leftarrow T_1$\;
    \For{$t \leftarrow 1, \ldots, s$}
        {
        Sample minibatch $B_t$ from $T_c$\;
        $\theta_{t+1} \leftarrow \theta_t - \eta\nabla_{\theta_{t}} L_{B_t}(\theta_t)$\;
        \If{\normalfont{changedTask}}
            {
            changedTask $\leftarrow$ False\;
            }
        \ElseIf {$D'_c(\theta_{t-1},\theta_{t}) < \alpha D'_c(\theta_{t-2},\theta_{t-1})$}
            {
            $C_c \leftarrow D'_c(\theta_{t-1},\theta_{t})$\;
            $T_c \leftarrow \text{sample}^{*}(\mathcal{T} - \{T_c\})$\;
            changedTask $\leftarrow$ True\;
            }
        $t \leftarrow t+1$\;
        }
    \caption{Self-paced scheduling algorithm for MNMT using the variation of model weights.}
    \label{alg:dynamic_sampling}
\end{algorithm}

In training, we use the `noam' learning rate schedule \citep[Eq.~3]{NIPS2017_3f5ee243}, which increases linearly from zero during the warmup steps, and afterwards decays proportionally to the inverse square root of the current step. Although the variation in weights throughout the entire training is strongly influenced by the learning rate schedule (Figure~\ref{fig:KL_div_smoothing}), we find that when comparing the smoothed weight variations between two near-consecutive steps, the influence of the learning rate variation is negligible. Finally, we note that our algorithm carries little computational overhead, since the self-assessed competence is obtained from the weight variation across standard updates.

\subsection{Explorations of our Method}

Firstly, we study the effect of several metrics to measure weight variation (Section~\ref{subsec:variation_metric}). Next, we perform various experiments in order to optimize our method, involving: the smoothing parameter $w$ (Section~\ref{subsec:smoothing}), the importance of the previous weight variation $\alpha$ (Section~\ref{subsec:alpha}), and training during the warmup steps only on the HRL, which simulates a pre-training regime (Section~\ref{subsec:hrl_warmup}). Experiments in Sections~\ref{subsec:smoothing} to~\ref{subsec:hrl_warmup} are performed in a 2-to-1 setup (\textsc{Gl}-\textsc{Pt}-to-\textsc{En}) using default hyper-parameters.

\subsubsection{Weight Variation Metric}
\label{subsec:variation_metric}

\begin{figure*}[ht!]
    \centering
    \includegraphics[scale=0.6]{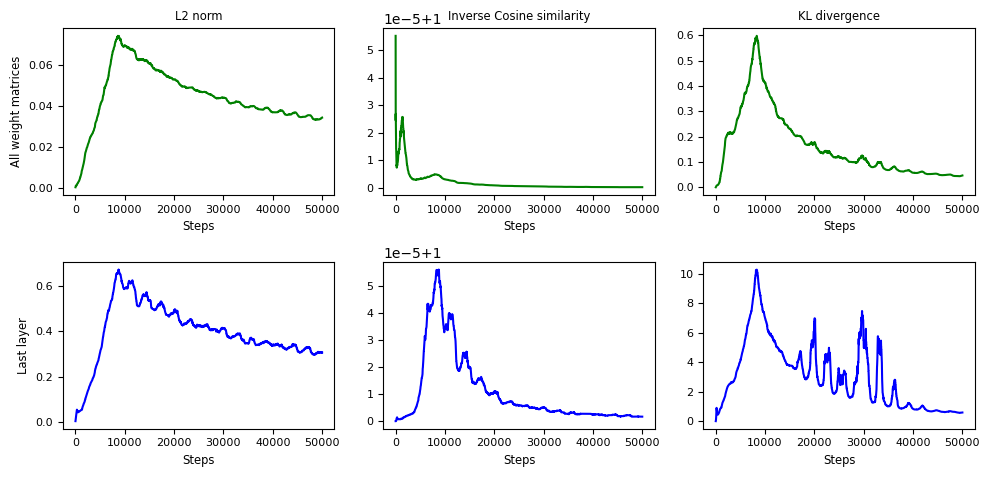}
    \caption{Comparison between three different metrics for model weight variation: L2 norm, inverse cosine similarity, and KL divergence. For each of them we compare monitoring the average over all weight matrices and only the final output layer.}
    \label{fig:measures_comparison}
\end{figure*}

In order to measure the weight variation of a model between steps, firstly we train a model on a unidirectional low-resource NMT task (60k lines) and compare measuring the average of all weight matrices versus the last output layer, and using KL divergence as our metric, inverse cosine similarity, or L2 norm. We show in Figure~\ref{fig:measures_comparison} these six combinations, computing the variations every 10 steps and performing a rolling average with a window size of 100. We can observe in all of them the effect of the learning rate schedule (warmup steps and decay). Additionally, we also note a more regular pattern when measuring the change across all matrices versus the last layer. We decide on using KL divergence as our measure due to it striking a balance between the irregularity of the inverse cosine similarity and the L2 norm.

\subsubsection{Setting of the Smoothing Weight}
\label{subsec:smoothing}

\begin{figure}[ht]
    \centering
    \includegraphics[width=0.49\textwidth]{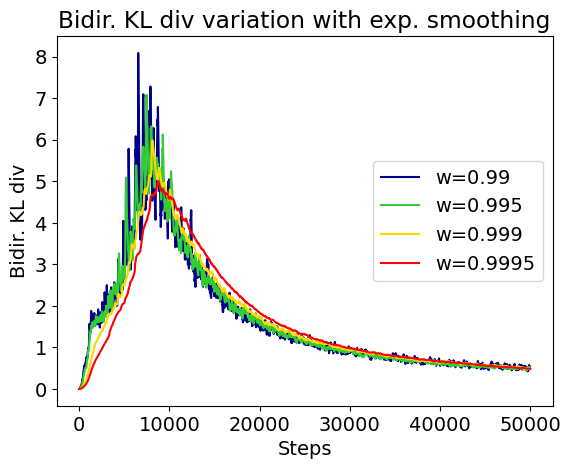}
    \caption{Evolution of the bidirectional Kullback-Leibler divergence for different values of the exponential smoothing coefficient $w$ in an experiment on \textsc{Gl}-\textsc{Pt}$\rightarrow$\textsc{En} with dynamic sampling}
    \label{fig:KL_div_smoothing}
\end{figure}

\begin{table}[ht!]
\centering
\begin{tabular}{c l c c c}
\toprule
% \small
\textbf{\#} & \textbf{System} & \textbf{$w$} & \textbf{Updates} & \textbf{\textsc{Bleu}}\\
\midrule
& \textit{shuffled} & - & 44k & \textbf{27.49}\\
& \textit{alternation} & - & 44k & 25.64\\
\hline
\rectangle{blue}& \textit{self-paced} & 0.99 & 48k & 25.76 \\
\rectangle{green}& \textit{\underline{self-paced}} & \underline{0.995} & \underline{48k} & \underline{25.92} \\
\rectangle{yellow}& \textit{self-paced} & 0.999 & 40k & \textbf{26.28} \\
\rectangle{red}& \textit{self-paced} & 0.9995 & 28k & 25.42 \\
\bottomrule
\end{tabular}
\caption{\textsc{Bleu} scores on the LRL test set of our method with several values of the smoothing coefficient, $w$. We denote in bold the best result in the comparison methods, as well as in our method, and we underline our chosen value for $w$.}
\label{tab:results-app-smoothing}
\end{table}

In Figure~\ref{fig:KL_div_smoothing} we show the average weight variation between all weight matrices when experimenting with various values for $w$, and in Table~\ref{tab:results-app-smoothing} the resulting scores, which we compare to \textit{shuffled} and \textit{alternation}. We can see that increasing $w$ not only produces a more regular weight-variation curve, but also accelerates training without much loss in test score. Nonetheless, although some of the smoothing values produce better scores than a simple \textit{alternation} of monolingual batches, none of them improve over the multilingual \textit{shuffled} batches. We select a $w=0.995$ for our main experiments as a balance between translation quality and regularity of the weight variation curve.

\subsubsection{Importance of Previous Weight Variation}
\label{subsec:alpha}

\begin{table}[ht]
\centering
\begin{tabular}{l c c c}
\toprule
\textbf{System} & \textbf{$\alpha$} & \textbf{Updates} & \textbf{\textsc{Bleu}} \\
\midrule
\textit{shuffled} & - & 44k & \textbf{27.49}\\
\textit{alternation} & - & 44k & 25.64\\
\hline
\textit{self-paced} & 0.9 & 48k & 11.12 \\
\textit{self-paced} & 0.95 & 48k & 11.91\\
\textit{\underline{self-paced}} & \underline{1.0} & \underline{48k} & \textbf{\underline{25.92}}\\
\textit{self-paced} & 1.1 & 40k & 25.04\\
\textit{self-paced} & 1.2 & 44k & 25.37\\
\bottomrule
\end{tabular}
\caption{\textsc{Bleu} scores on the LRL test set of our method with several values of the importance hyper-parameter $\alpha$, with $w=0.995$. We denote in bold the best result in the comparison methods, as well as in our method, and we underline our chosen value for $\alpha$.}
\label{tab:results-app-alpha}
\end{table}

Similarly, we also experiment on the value of $\alpha$, a hyper-parameter to weight the importance of the previous weight variation when comparing steps $t$ and $t-1$. In Table~\ref{tab:results-app-alpha} we show the results of training with our selected value of $w=0.995$ and various values for $\alpha$. We observe best results without any additional weighting of the previous variation, and so select for our main experiments $\alpha = 1$.

\begin{figure*}[ht]
    \centering
    \includegraphics[width=0.98\textwidth]{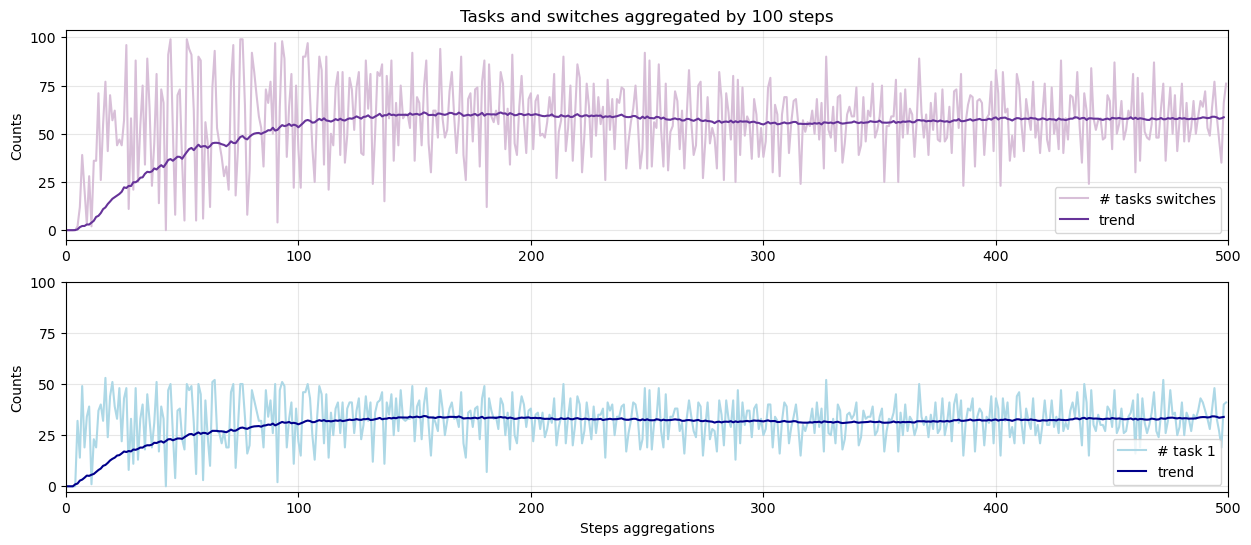}
    \caption{Amount of task switches and percentage of training on the LRL (\textit{task 1}).}
    \label{fig:task_switches}
\end{figure*}

\subsubsection{Training with HRL Warmup Steps}
\label{subsec:hrl_warmup}

\begin{table}[ht]
\centering
\begin{tabular}{l c c c}
\toprule
\multirow{2}{*}{\textbf{System}} & \textbf{HRL} & \multirow{2}{*}{\textbf{Updates}} & \multirow{2}{*}{\textbf{\textsc{Bleu}}} \\
& \textbf{warmup} & & \\
\midrule
\textit{alternation} & no & 36k & 24.92 \\
\textit{alternation} & yes & 44k & \textbf{26.04} \\
\midrule
\textit{\underline{self-paced}} & \underline{no} & \underline{48k} & \underline{25.92} \\
\textit{self-paced} & yes & 48k & \textbf{26.16} \\
\bottomrule
\end{tabular}
\caption{\textsc{Bleu} scores on the LRL test set of our method when observing the role of HRL warmup, with $\alpha=1$ and $w=0.995$. We denote in bold the best result in the comparison methods, as well as in our method, and underline our chosen final technique.}
\label{tab:results-app-warmup}
\end{table}

Due to the effect that the learning rate warmup steps has on weight variation during the first 8k steps of training, we also consider starting training the two methods involving monolingual batches (\textit{alternation} and \textit{self-paced}) exclusively on the HRL, which simulates a pre-training regime. We show in Table~\ref{tab:results-app-warmup} the effects of HRL warmup between each of these two methods. We observe \textit{alternation} benefits significantly (+1 \textsc{Bleu} points) from HRL warmup, but our \textit{self-paced} method much less noticeably. We do not consider HRL warmup to produce a positive balance between complexity and score improvement for our method, so we do not perform it in our main experiments.

\subsubsection{Amount of Task Switches and Balancing}
\label{subsec:task_switches}

Finally, in Figure~\ref{fig:task_switches} we show the amount of task switches in training, aggregated by 100 steps, where \textit{task 1} is the LRL. We can see that on this 2-to-1 case, after initial learning rate warmup steps, our method settles on a third of the training consisting of the LRL and two thirds on the HRL.

% = = = = = = = = = = = = = = = = = = = = = = = = = = =
\section{Data and Systems}
\label{sec:data-system}

\subsection{Corpora}
\label{subsec:corpora}

We experiment on a subset of the multilingual TED corpus \citep{qi-etal-2018-pre}. As in previous multilingual studies \citep{neubig-hu-2018-rapid,wang2019multilingual}, we focus on four pairs of related LRL-HRL, as shown in Table~\ref{tab:data}, with the goal of translating them into English (\textsc{En}).

\begin{table}[ht]
	\centering
        % \small
	\begin{tabular}{A r r r A r}
		\toprule
            \textbf{LRL} & \texttt{train} & \texttt{dev} & \texttt{test} & \textbf{HRL} & \texttt{train} \\
            \cmidrule(lr){1-4} \cmidrule(lr){5-6}
  		\textsc{Be} & 4.51k          & 248          & 664 & \textsc{Ru} & 208k            \\
            \textsc{Az} & 5.94k          & 671          & 903 & \textsc{Tr} & 182k            \\
            \textsc{Gl} & 10.0k          & 682          & 1.0k & \textsc{Pt} & 51.8k            \\
            \textsc{Sk} & 61.5k          & 2.2k        & 2.4k & \textsc{Cs} & 103k            \\
		\bottomrule
	\end{tabular}
	\caption{Data sizes for pairs of LRLs and HRLs.}
	\label{tab:data}
\end{table}

\subsection{Tokenization}
\label{subsec:subword_tokenization}

As the data is already tokenized, we directly use Byte Pair Encoding (BPE) \citep{sennrich-etal-2016-neural} for subword extraction and vocabulary construction. We learn a vocabulary by concatenating 10k random lines from each language in the training data, and upsample the LRL if it has fewer lines. For experiments involving only one LRL and one HRL in the source, we learn a vocabulary of 10k subwords.  For experiments involving all four LRLs and all four HRL in the source, our vocabulary has 32k subwords.  To facilitate language identification, we prefix the dataset of each language with a unique tag.

\subsection{System Architecture}
\label{subsec:architecture}

We use Transformer models \citep{NIPS2017_3f5ee243} from the OpenNMT-py library \citep{klein-etal-2017-opennmt} version 3.1.1. In all our systems we use the following default values of hyper-parameters from Transformer-Base: 6 encoder/decoder layers, 8 attention heads, label smoothing of 0.1, hidden layer of 512 units, and FFN of 2,048 units. We use Adam optimizer \citep{kingma-adam2014} and a batch size of 10k tokens. 

Moreover, we compare systems trained with default regularization to systems using more aggressive regularization (see Section~\ref{sec:results}). The former consist of a dropout rate of 0.1, OpenNMT-py's scaling factor of 2 over the learning rate, 8k warmup steps, and no gradient clipping. For the latter, we increase the dropout rate to 0.3, the scaling factor to 10 and the number of warmup steps to 16k, and we re-normalize gradients if their norm is greater than 5.

\subsection{Evaluation}
\label{subsec:evaluation}

For each pair, we measure the \textsc{Bleu} score \citep{papineni-etal-2002-bleu} on the LRL test set using the SacreBLEU library\footnote{\href{https://github.com/mjpost/sacrebleu}{github.com/mjpost/sacrebleu}\\signature: \texttt{nrefs:1|case:mixed|eff:no|tok:13a |smooth:exp|version:2.3.1.}} \citep{post-2018-call} as well as the COMET score \citep{rei-etal-2020-comet} using model \texttt{wmt22-comet-da}. We use bootstrap resampling from SacreBLEU to compute the 95\% confidence interval around the mean of the \textsc{Bleu} score.  We use a rolling ensemble of four checkpoints and select the best on the development set for the final translations.

% = = = = = = = = = = = = = = = = = = = = = = = = = = =
\section{Results}
\label{sec:results}

We search first for the appropriate level of regularization to apply to our approach, by considering 2-to-1 MNMT systems (four systems, HRL and LRL to \textsc{En}, with data shown in Table~\ref{tab:data}). For each MNMT system, we compare three methods: first, we train a model on multilingual batches, by upsampling all the tasks until they are the same size and then \textit{shuffling} them.  Second, we apply a cyclical \textit{alternation} of monolingual batches for each task, which results in the model being trained the same amount of time on all tasks.  Third, we apply our \textit{self-paced method} described in Section~\ref{sec:self-paced}. For each of the methods we compare a model trained with default hyper-parameters and more regularized ones \citep{atrio-popescu-belis-2022-interaction}.  

The average \textsc{Bleu} scores over the four LRL tests sets of each model are shown in Table~\ref{tab:results-main-regularization}. Training with more regularization improves all three methods by 3 to 3.5 \textsc{Bleu} points. Additionally, the more regularized models improve over the scores of previous studies on the same data \citep{neubig-hu-2018-rapid,aharoni-etal-2019-massively,wang-etal-2020-balancing}. On this 2-to-1 setup we obtain better results when training with multilingual \textit{shuffled} batches, and a small improvement of \textit{self-paced} versus an \textit{alternation} of monolingual batches.
All regularized models methods reach their highest \textsc{Bleu} score at a very similar number of updates, although when training with more aggressive regularization, we observe a noticeable improvement in speed in the \textit{self-paced} method.

\begin{table}[htb] % ARA: neubig with LSTMS has a 21.9, wang a 22.4, and aharoni (with plenty more languages) a 23.7 BLEU.
\centering
% \small
\begin{tabular}{l r r r}
    \toprule
    \textbf{System} & \textbf{\textsc{Bleu}} & \textbf{COMET} & \textbf{Updates} \\
    \cmidrule(lr){1-1} \cmidrule(lr){2-4}
    \textit{shuffled} & 22.1 & 64.7 & 36 \\
    ~+ regularized & \textbf{25.3} & \textbf{69.2} & \textbf{39} \\
    \cmidrule(lr){1-1} \cmidrule(lr){2-4}
    \textit{alternation} & 20.9 & 63.2 & 35 \\
    ~+ regularized & \textbf{24.4} & \textbf{68.0} & \textbf{40} \\
    \cmidrule(lr){1-1} \cmidrule(lr){2-4}
    \textit{self-paced} & 21.2 & 63.8 & 49 \\
    ~+ regularized & \textbf{24.6} & \textbf{68.5} & \textbf{36} \\
    \bottomrule
\end{tabular}
\caption{Average \textsc{Bleu} scores on the test sets of the four LRLs on 2-to-1 setups for the three sampling strategies, with standard and increased regularization
(best scores in bold). We also present the amount of updates required to obtain each best score, in thousands.}
\label{tab:results-main-regularization}
\end{table}

In Table~\ref{tab:results-main-two-eight} we present the scores of models trained on the four 2-to-1 setups from Table~\ref{tab:results-main-regularization} by order of increasing LRL size, as well as an 8-to-1 setup, which includes all the tasks. Introducing more source languages does not improve scores in either of the three methods, although we observe a small negative effect on higher-resourced LRLs, which has been reported previously \citep{neubig-hu-2018-rapid,aharoni-etal-2019-massively}.
\textit{Self-paced} tends to perform better than \textit{alternation} on 2-to-1, but is more severely affected on an 8-to-1 setup. Both of these methods, which rely on monolingual updates, clearly underperform with respect to \textit{shuffled}, which is trained with multilingual batches.

\begin{table*}[ht!] % ARA: if we move the table before for the layout, the numbering doesn't make as much sense (https://tex.stackexchange.com/questions/2727/forcing-specific-numbers-for-tables ?)
	\centering
	\begin{tabular}{llcccccc}
        \toprule
        
        \multirow{2}{*}{\textbf{Language}} & \multirow{2}{*}{\textbf{System}} & \multicolumn{3}{c}{\textbf{2-to-1}} & \multicolumn{3}{c}{\textbf{8-to-1}} \\
                           & & \textbf{\textsc{Bleu}}               & \textbf{COMET} & \textbf{Updates} & \textbf{\textsc{Bleu}}   & \textbf{COMET} & \textbf{Updates} \\
        % \midrule
        \cmidrule(lr){1-2} \cmidrule(lr){3-5} \cmidrule(lr){6-8}
        \multirow{3}{*}{\textsc{Be}$\rightarrow$ \textsc{En}} & \textit{shuffled}        & \textbf{21.7} \footnotesize{$(\pm 1.3)$} & \textbf{63.8} & \textbf{48k} & \textbf{20.0} \footnotesize{$(\pm 1.4)$} & \textbf{61.4} & \textbf{64k} \\
                           & \textit{alternation}     & 19.8 \footnotesize{$(\pm 1.2)$}          & 61.5          & 44k & 18.7 \footnotesize{$(\pm 1.3)$}          & 61.3 & 120k           \\
                           & \textit{self-paced}         & 20.5 \footnotesize{$(\pm 1.3)$}          & 62.8          & 32k & 19.7 \footnotesize{$(\pm 1.3)$}          & 61.8 & 128k \\
        % \hline
        \cmidrule(lr){1-2} \cmidrule(lr){3-5} \cmidrule(lr){6-8}
        \multirow{3}{*}{\textsc{Az}$\rightarrow$ \textsc{En}} & \textit{shuffled}        & \textbf{15.6} \footnotesize{$(\pm 1.0)$} & \textbf{66.0} & \textbf{32k} & 14.3 \footnotesize{$(\pm 1.0)$}          & 64.4 & 144k \\
                           & \textit{alternation}     & 14.4 \footnotesize{$(\pm 1.0)$}          & 63.9          & 44k & \textbf{16.6} \footnotesize{$(\pm 1.0)$}          & \textbf{62.9}         & \textbf{140k} \\
                           & \textit{self-paced}         & 14.5 \footnotesize{$(\pm 1.0)$}          & 64.9          & 48k & 14.4 \footnotesize{$(\pm 1.0)$} & 64.0          & 150k\\
        % \hline
        \cmidrule(lr){1-2} \cmidrule(lr){3-5} \cmidrule(lr){6-8}
        \multirow{3}{*}{\textsc{Gl}$\rightarrow$ \textsc{En}} & \textit{shuffled}        & \textbf{30.2} \footnotesize{$(\pm 1.2)$}          & \textbf{70.9}          & \textbf{50k} & \textbf{31.9} \footnotesize{$(\pm 1.3)$} & \textbf{72.8} & \textbf{92k} \\
                           & \textit{alternation}     & 30.0 \footnotesize{$(\pm 1.2)$}          & 71.0         & 50k & 30.4 \footnotesize{$(\pm 1.2)$}           & 71.3         & 136k \\
                           & \textit{self-paced}         & \textbf{30.2} \footnotesize{$(\pm 1.1)$} & \textbf{71.2} & \textbf{44k} & 30.7 \footnotesize{$(\pm 1.2)$}          & 72.0         & 150k \\
        % \hline
        \cmidrule(lr){1-2} \cmidrule(lr){3-5} \cmidrule(lr){6-8}
        \multirow{3}{*}{\textsc{Sk}$\rightarrow$ \textsc{En}} & \textit{shuffled}        & \textbf{33.6} \footnotesize{$(\pm 0.8)$} & \textbf{76.2} & \textbf{24k} & \textbf{33.9} \footnotesize{$(\pm 0.9)$} & \textbf{75.4} & \textbf{32k} \\
                           & \textit{alternation}     & 33.4 \footnotesize{$(\pm 0.8)$}          & 75.3          & 20k & 31.8 \footnotesize{$(\pm 0.9)$}          & 73.5  & 144k        \\
                           & \textit{self-paced}         & 33.3 \footnotesize{$(\pm 0.9)$}          & 75.3          & 20k & 31.9 \footnotesize{$(\pm 0.9)$}          & 74.0      & 128k    \\     
        % \hdashline
        \cmidrule[1.5pt](lr){1-2} \cmidrule[1.5pt](lr){3-5} \cmidrule[1.5pt](lr){6-8}
        
        \multirow{3}{2cm}{\textit{Average of the four LRLs}} & \textit{shuffled} & \textbf{25.3} & \textbf{69.2} & \textbf{39k} & \textbf{25.0} & \textbf{69.0} & \textbf{83k} \\
        & \textit{alternation} & 24.4 & 68.0 & 40k & 24.4 & 67.3 & 135k \\
        & \textit{self-paced} & 24.6 & 69.0 & 36k & 24.2 & 68 & 139k \\
        
        \cmidrule[1.5pt](lr){1-8}
	\end{tabular}
	\caption{Results of the three methods compared on four 2-to-1 setups and an 8-to-1 setup, as well as the number of updates necessary to obtain the highest scores. We use our stronger regularization hyper-parameters (as in Table~\ref{tab:results-main-regularization}), and denote in bold the method that obtains the best \textsc{Bleu} score for each task.}
	\label{tab:results-main-two-eight}
\end{table*}

Additionally, in the 2-to-1 case, the difference between \textit{shuffled} and \textit{self-paced} decreases as the size of the dataset increases, but in the 8-to-1 case, the difference increases as the dataset size also increases. % (1.2, 1.1, 0.0, 0.3 \textsc{Bleu} points) and (0.3, -0.1, 1.2, 2) respectively.
This indicates that with a small amount of training tasks, the more available data, the less important the sampling method is, but with many training tasks a more careful selection of the balancing of the data becomes more important for lower-resourced datasets.

Regarding convergence speed, we measure the amount of updates that each model requires in order to reach its highest \textsc{Bleu} scores. We observe that all three methods train in nearly the same speed on the 2-to-1 case, but on the 8-to-1 case we observe that training with monolingual batches, regardless of the balancing of the tasks, results in a much slower training. This is likely due to either the model forgetting what it learned the last time it trained on a given task, or to the monolingual updates resulting in weights that are less useful to the other tasks.

% = = = = = = = = = = = = = = = = = = = = = = = = = = =
\section{Conclusion and Future Work}
\label{sec:conclusion}

In this study we have presented a self-paced method to balance tasks in a many-to-one MNMT system by monitoring the average per-task weight variation across steps, with the objective of not over-training on tasks in which the model is competent, and better allocating resources to tasks in which the model is less competent. Our method carries no dedicated computational overhead. However, we have observed that a multilingual, uniform balancing of all tasks outperforms our method both on 2-to-1 and 8-to-1 setups.

A limitation of our method may be that measuring the weight variation between two consecutive updates might result in too small a value, with too many oscillations even with the use of smoothing. Additionally, we have shown that as the amount of training tasks increases, performing single-task updates is counter-productive, both in quality and in speed. In the future, we hope to extend our method to assemble multilingual batches based on the per-task weight variation in order to solve this issue.

% = = = = = = = = = = = = = = = = = = = = = = = = = = =
\section*{Ethics Statement}
% From ARR's page: “Authors are encouraged to devote a section of their paper to concerns about the ethical impact of the work and to a discussion of broader impacts of the work, which will be taken into account in the review process. This discussion may extend into a 5th page (short papers).”

This study does not process personal or sensitive data.  While MT in general may facilitate disclosure or cross-referencing of personal information, which may pose threats to minorities, the community appears to consider that the potential benefits far outweigh the risks, judging from the large number of studies for low-resource and unsupervised~MT.

% = = = = = = = = = = = = = = = = = = = = = = = = = = =
\section*{Acknowledgments}

We are grateful for the support received from Armasuisse (UNISUB projet: Unsupervised NMT with Innovative Multilingual Subword Models) and from the Swiss National Science Foundation (DOMAT project: On-demand Knowledge for Document-level Machine Translation, n.\ 175693).

\bibliography{anthology,custom}
\bibliographystyle{acl_natbib}

% = = = = = = = = = = = = = = = = = = = = = = = = = = =

\end{document}